\documentclass{article}

\usepackage{arxiv}

\usepackage{hyperref}
\usepackage[T1]{fontenc}
\usepackage{graphicx}
\usepackage{amssymb,amsmath,amsthm}
\usepackage{mathtools}
\usepackage{multirow}
\usepackage{booktabs}	
\usepackage{float}	
\usepackage{amsmath,amsfonts,latexsym,amssymb}
\usepackage[mathscr]{euscript}
\usepackage{mathtools}
\usepackage{subfigure}
\usepackage{cite}
\usepackage{enumitem}
\usepackage{bm}
\usepackage{color,soul} 
\usepackage{array}
\usepackage[ruled,linesnumbered]{algorithm2e}
\SetKwComment{Comment}{$\triangleright$\ }{}
\usepackage{siunitx} 
\usepackage{cancel} 
\usepackage{comment} 

\usepackage{diagbox}

\newlength{\figwidth}
\newlength{\subfigwidth}
\newlength{\subfigheight}
\renewcommand{\equationautorefname}{Eq.}

\newcolumntype{d}[1]{D{.}{.}{#1}}

\def\equationautorefname~#1\null{%
  Eq.~(#1)\null
  }
  
\title{Transfer learning based multi-fidelity physics informed deep neural network}

\author{
 Souvik Chakraborty \\
  Department of Applied Mechanics,\\
  Indian Institute of Technology Delhi,\\
  New Delhi, India.\\
  \texttt{csouvik41@gmail.com} \\
   }
 
\begin{document}
\maketitle
\begin{abstract}
For many systems in science and engineering, the governing differential equation is either not known or known in an approximate sense.
Analyses and design of such systems are governed by data collected from the field and/or laboratory experiments.
This challenging scenario is further worsened when data-collection is expensive and time-consuming.
To address this issue, this paper presents a novel 
multi-fidelity physics informed deep neural network
(MF-PIDNN).
The framework proposed is particularly suitable 
when the physics of the problem is known in an approximate sense (low-fidelity physics) and only a few high-fidelity data are available.
MF-PIDNN blends physics informed and data-driven deep learning techniques by using the concept of 
transfer learning.
The approximate governing equation is first used
to train a low-fidelity physics informed deep neural
network.
This is followed by transfer learning where the 
low-fidelity model is updated by using the available
high-fidelity data.
MF-PIDNN is able to 
encode useful information on the physics 
of the problem from the {\it approximate} governing differential equation 
and hence, provides accurate prediction even in zones with no data. Additionally, no low-fidelity data is required for training this model.
Applicability and utility of MF-PIDNN are illustrated in solving four benchmark reliability analysis 
problems. Case studies to illustrate interesting features of the proposed approach are also presented.
\end{abstract}


	
\section{Introduction}
\label{sec:intro}
The governing equations used in science and engineering are often 
based on certain assumptions and approximations \cite{brunton2016discovering}.
For example, heterogeneous material properties
are approximated as homogeneous \cite{rodrigues2018material}, effect
of environmental conditions are rarely considered \cite{haapio2008environmental}
and critical parts such as 
joints are often ignored \cite{mandal2015simple}.
Naturally, results obtained by solving the governing
equations only provide an approximation of the true system behavior (i.e., low-fidelity results).
An alternative is to perform actual experiments
in a laboratory environment.
With modern experimental setups and sensors, 
it is possible to perform highly sophisticated 
experiments \cite{ramesh2016digital,mennel2020ultrafast}. Results obtained from such experiments
are generally accurate (high-fidelity results).
However, experiments
are expensive and time-consuming, and
one can only perform a limited number of experiments (usually in the order of tens).
Such small number of experiments is often
not sufficient for understanding the system behavior,
specifically if dealing with problems such
as uncertainty quantification and reliability analysis.

One possible solution to the difficulties raised above
resides in multi-fidelity schemes \cite{peherstorfer2018survey,giselle2019issues,chakraborty2017surrogate} where
data fusion techniques are used to combine 
high-fidelity and low-fidelity data.
The most popular multi-fidelity schemes are  perhaps
the multi-level Monte Carlo (MLMC) methods \cite{bierig2016approximation,giles2008multilevel,giles2017adaptive,heinrich2001multilevel}.
The primary idea in MLMC is to accelerate the calculation of the second moments of the quantity of
interests.
Another popular approach for dealing with multi-fidelity
data is co-Kriging \cite{le2014recursive,perdikaris2015multi,koziel2012variable,le2013multi}.
In this method, Kriging \cite{biswas2017kriging,mukhopadhyay2017critical,saha2018kriging,kaymaz2005application}, aka Gaussian process \cite{Bilionis2012multi,Bilionis2013multi,Nayek2019a,Chakraborty2019graph,chakraborty2020role}, 
is coupled with an auto-regressive like
information fusion scheme \cite{babaee2016multi,batra2019multifidelity,perdikaris2016multifidelity}.
Methods where the Gaussian process in co-Kriging
is replaced by other machine learning techniques
can also be found in the literature \cite{liu2016multi,park2017remarks,yan2019adaptive,palar2016multi}.
The success of all these methods is already well-established in
the literature \cite{gao2020bi,forrester2008engineering}.
Unfortunately, these methods only work for cases where
the low-fidelity data is able to capture the trend
and the models of different fidelities have a strong
linear correlation 
Both co-Kriging motivated approaches and
MLMC fails when the low-fidelity and high-fidelity
data have a space-dependent, complex and
nonlinear correlations.
To address this issue, researchers have recently proposed methods that are rooted in  Bayesian statistics \cite{nitzler2020generalized} and nonlinear auto-regressive algorithm \cite{perdikaris2017nonlinear}.

The field of artificial intelligence and machine 
learning has recently witnessed a huge boom \cite{goodfellow2016deep} and
its influence can also be observed in the 
multi-fidelity approaches.
De {\it et al.} \cite{de2020transfer} developed two multi-fidelity approaches by using deep neural networks.
While the first framework uses transfer learning, 
the second framework utilizes bi-fidelity weighted
learning.
Meng and Karniadakis \cite{meng2020composite}, on the other hand,
proposed a composite neural network that 
is trained based on multi-fidelity data.
A physics aware component was also added to this
network; although, the physics informed 
component is only used for solving inverse problems.
Liu and Wang \cite{liu2019multi} proposed physics
constrained multi-fidelity neural networks for solving partial differential equations.

Based on the discussion above,
(at least) two salient conclusions can be drawn about the existing multi-fidelity approaches.
\begin{itemize}
    \item First, the existing multi-fidelity approaches
assume the low-fidelity solver to be computationally efficient so that one can generate
sufficient low-fidelity data.
This is not always true. 
For example, compared to wind tunnel test data, 
a large eddy simulation \cite{zhiyin2015large,barone2017machine} solver can be treated as 
a low-fidelity solver.
However, computational cost associated with large eddy simulation is significant, even on modern computer clusters.

\item Second, the physics informed multi-fidelity
approaches proposed in \cite{liu2019multi} assume that the
{\it exact} physics corresponding to the high-fidelity data is known. This is not necessarily true. 
There are problems where the underlying physics is unknown \cite{brunton2016discovering}.
Also, the apparently known governing equations are often
derived based on certain assumptions and hence, only
reflect the true scenario in an approximate manner.
\end{itemize}

The objective of this paper is to present a multi-fidelity physics informed deep learning 
framework that addresses both the limitations discussed above.
Unlike some of the previous studies, it is assumed that the data-generation process for the high-fidelity 
data is unknown. 
The low-fidelity model is given
by ordinary/partial differential equations.
The proposed model needs no low-fidelity data; instead, the initial low fidelity model is directly trained
based on the (approximate) physics of the problem. 
This is achieved by utilizing the recently 
developed physics informed deep learning algorithm.
\cite{raissi2019physics,goswami2020transfer,chakraborty2020simulation,zhu2019physics}.
With this setup, important physical laws such as 
invariance and symmetries present in the low-fidelity model will be inherently captured by
the deep learning framework.
Transfer learning \cite{goswami2020transfer} and available high-fidelity data is then used to update the trained deep learning framework. 
Performance of the proposed framework is illustrated on selected reliability analysis problems from the literature.

The rest of the paper is organized as follows.
\autoref{sec:ps} provides details on the problem to be solved.
Details about the proposed approach are presented in
\autoref{sec:pa}.
Numerical results showcasing the performance of the proposed approach are presented in \autoref{sec:ni}.
Finally, \autoref{sec:conclusions} provides the concluding remarks.
\section{Problem statement}
\label{sec:ps}
Consider $\bm \Xi = \left( \Xi_1, \Xi_2,\ldots, \Xi_N \right): \Omega \rightarrow \mathbb R^N$ to be an $N-$dimensional stochastic vector with cumulative distribution function 
\begin{equation}\label{eq:cdf}
    F_{\bm \Xi}(\bm \xi) = \mathbb P\left( \bm \Xi \le \bm \xi \right),    
\end{equation}
where $\bm \xi$ is a realization from the random vector $\bm \Xi$, $\mathbb P \left(\cdot\right)$
represents the probability measure and $\Omega$ is 
the input domain.
In reliability analysis, one first formulates a 
limit-state or performance function, $\mathcal J \left( \bm \xi \right) = 0$ such that $\mathcal J\left( \bm \xi \right) < 0$ represents the failure
domain $\left( \Omega_f \right)$ and $J\left( \bm \xi \right) \ge 0$ represents
the safe domain.
Mathematically, this can be represented as
\begin{equation}\label{eq:failure_domain}
    \Omega_f \triangleq \left\{ \Xi: \mathcal J \left( \bm  \xi < 0 \right) \right\}.
\end{equation}
With this consideration, the probability of failure
of the system can be calculated as
\begin{equation}\label{eq:pf}
\begin{split}
    P_f = \mathbb P \left( \bm \Xi \in \Omega_f \right) & = \int_{\Omega_f}{\text d F_{\bm \Xi}\left( \bm \xi \right)}\\
    & = \int_{\Omega}{\mathbb I_{\Omega_f}\text d F_{\bm \Xi}\left( \bm \xi \right)},
\end{split}
\end{equation}
where $\mathbb I_{c}$ is an indicator function,
\begin{equation}\label{eq:indicator}
    \mathbb I_{c}\left( \bm \xi \right) = \left\{ 
    \begin{array}{lcl}
         1 & \text{ if} & \bm \xi \in c \\
         0 & \text{ if} & \bm \xi \notin c
    \end{array}
    \right.
\end{equation}
Although the mathematical formulation of reliability
analysis discussed above is relatively simple, the 
difficulty arises due to the multivariate integral
in \autoref{eq:pf}.
Almost all the time, there exists no closed-form 
solution for the multivariate integral and hence, one
has to rely on numerical integration techniques or asymptotic approximations. 
A detailed account of different reliability analysis methods can be found in \cite{haldar2000probability,haldar2000reliability}.

Another important player in reliability analysis is
the limit-state function $\mathcal J \left(\bm \xi \right)$.
For computing the probability of failure in
\autoref{eq:pf} using numerical integration, 
one needs to evaluate the limit-state function
repeatedly;
the inherent assumption being, the mathematical
model/equation for the limit-state function is 
known.
In this regards, the accuracy of the limit-state function
is of utmost importance.
However, for many systems in science and engineering,
the governing equation is either not available or available in an approximate sense \cite{brunton2016discovering}.
Under such circumstances, one has no option but
to rely on data collected either from the field or 
from laboratory experiments.
Further assuming that the system under consideration is at its
design phase, the option of collecting field data 
becomes invalid and performing laboratory experiments
is the only feasible alternative.

Consider, $\mathcal D_{h} = \left[ \bm \Xi_{hx}, \bm u_h \right]$ to be the data available from laboratory experiments where
\begin{equation}\label{eq:hf_data_input}
    \bm \Xi_{hx} =  \bm \Xi_h \otimes \bm x_h \otimes \bm t_h.
\end{equation}
$\bm \Xi_h = \left[\bm \xi^{(1)}, \ldots, \bm \xi^{(N_{h})} \right]^T$ in \autoref{eq:hf_data_input} represents
sample/data of the stochastic inputs,
$\bm x_h = \left[x_1,\ldots,x_s  \right]^T$ are the spatial locations where data is available (sensor locations) and $\bm t_h = \left[t_1,\ldots,t_n \right]^T$ are the time-steps are the times at which 
observations are available.
The operator `$\otimes$' in \autoref{eq:hf_data_input}
indicates Kronecker product and $\bm u_h = \left[u_1,\ldots u_{r}\right],\; r = N_{h}\times s \times n$ represents
the responses.
`$h$' in the suffix indicates that the data-collected
is high-fidelity.
The limit-state function $\mathcal J \left(\bm \xi \right)$ is generally expressed in terms of the response variable $u$ and a threshold $u_t$
\begin{equation}\label{eq:ls2}
    \mathcal J\left(\bm \xi \right) = g\left( u(\bm \xi, x_i, t_j) \right) - u_{t}.
\end{equation}
In case the number of data-points $N_{h}$ is significant,
it is possible to directly train a surrogate model, $\mathcal M: \bm (\bm \xi, x, t)  \rightarrow u$ and
then use it to evaluate the 
probability of failure in \autoref{eq:pf}.
Popular surrogate models available in the literature
includes Gaussian process \cite{Bilionis2013multi,Bilionis2012multi}, polynomial chaos expansion \cite{Xiu2002the,Sudret2008global}, analysis of variance decomposition \cite{Chakraborty2016sequential,Chakraborty2017polynomial}, support vector machine \cite{roy2020support} and hybrid polynomial correlated function expansion \cite{chakraborty2017efficient,chakraborty2017hybrid}.
However, in reality, the number of laboratory experiments that can be performed is limited and 
hence, the number of data-points available is 
often not sufficient for training a surrogate model.
To compensate for the fact that only a limited 
number of high-fidelity data is available, 
the approximate (low-fidelity) governing equation of the system is considered,
\begin{equation}\label{eq:lf_gde}
    u_t + h\left(u,u_x,u_{xx},\ldots;\bm \xi \right) = 0.
\end{equation}
$u_x$ and $u_{xx}$ in \autoref{eq:lf_gde} represent
the first and second derivative of $u$ with respect to $x$.
As already discussed in \autoref{sec:intro},
solving \autoref{eq:lf_gde} to generate sufficient number of 
low-fidelity data can also be computationally
expensive.

The objective of this paper is to develop a 
multi-fidelity deep learning framework that
can be directly trained by using the low-fidelity model in \autoref{eq:lf_gde} (without generating data from it)
and the high-fidelity data, $\mathcal D_{h}$.
\section{Multi-fidelity physics informed deep neural network}
\label{sec:pa}
In this section, the proposed multi-fidelity physics informed deep neural network (MF-PIDNN) is presented.
However, before proceeding to the proposed framework,
details on data-driven and physics-informed deep neural networks are discussed.
Data-driven and physics informed deep neural networks form
the backbone of the proposed multi-fidelity approach.
\subsection{Data-driven deep neural networks}
\label{subsec:ddnn}
One of the primary components of the proposed 
multi-fidelity approach is a deep neural network (DNN).
In this work, a fully connected DNN (FC-DNN) is used
and hence, the discussion is limited to FC-DNN only.
Having said that, the framework presented is generic
and can be used with convolutional \cite{zhu2018bayesian}
and other types of neural networks as well.

An FC-DNN with $L-$hidden layers can be represented
by using a sequence of activation functions and 
linear transformations
\begin{equation}\label{eq:fc-dnn}
    \mathbb N\left(\cdot ; \bm \theta \right) = \left( \sigma_{L} \circ \mathbf W_{L+1} \right)\circ \cdots \circ \left( \sigma_0 \circ \mathbf W_1 \right),
\end{equation}
where $\sigma_j: \mathbb R \rightarrow \mathbb R$ and $\mathbf W_{j+1}$ respectively represents the 
activation function and the weight matrix associated
with the edges connecting the $j-$th and $(j+1)-$th 
layers.
The biases of the neural network are absorbed
into the weight matrix $\mathbf W_j$; the weight matrices $\left\{\mathbf W_j\right\}_{j=1}^{L+1}$
are the parameters of the FC-DNN and are represented
using $\bm \theta$.
`$\circ$' in \autoref{eq:fc-dnn} represents operator
composition.
Note that the $0-$th layer in \autoref{eq:fc-dnn} represents the input and $(L+1)-$th layer
represents the output.
For using the neural network in practice, the model
parameters $\bm \theta$ needs to be estimated.
In a data-driven setting, this is achieved by 
minimizing a loss function.
For a detailed account of different loss-functions
available in the literature, interested readers may
refer \cite{goodfellow2016deep,beale1996neural}.
In this work, the mean-square loss function $\left( \mathcal L_d\right)$ has been used,
\begin{equation}\label{eq:loss}
    \mathcal L_d = \frac{1}{N_d}\sum_{k=1}^{N_d}\left(u_k - \hat u_k  \right)^2.
\end{equation}
In \autoref{eq:loss}, $N_d$ represents the number of 
data-points, $u_k$ is the observed response corresponding to the $k-$th input, $\bm \xi_k$ and
$\hat u_k$ represents the neural network predicted
response corresponding to $\bm \xi_k$,
\begin{equation}\label{eq:pred}
    \hat u_k = \mathbb N(\bm \xi_k;\bm \theta).
\end{equation}

The primary challenge behind the application of the DNN
for engineering applications is the need for data.
It is a well-acknowledged fact that DNNs are data-hungry tools \cite{nitzler2020generalized}.
Unfortunately, for the current work, the focus is on
problems where one has access to very few high-fidelity data.
Therefore, the direct application of data-driven DNN 
is unlikely to yield satisfactory results.
\subsection{Physics-informed deep neural networks}
\label{subsec:pinn}
To address the over-reliance of data-driven DNNs on
training data, physics informed deep neural 
networks (PI-DNN) was proposed in \cite{raissi2019physics}.
The basic idea is to compute the DNN 
parameters directly from the physics (governing ODE/PDE) of the problem.
Since its inception, the PI-DNN has been used for
solving a wide range of problems in science and engineering \cite{zhu2019physics,chakraborty2020simulation,goswami2020transfer,meng2020composite}.

Consider the governing (stochastic) differential equation
in \autoref{eq:lf_gde}. 
The objective is to solve the stochastic differential
equation so as to build a mapping from the input space (stochastic, spatial and temporal inputs) to
the response space.
In conventional data-driven DNN, this is achieved in three simple steps 
\begin{itemize}
    \item Generate training data $\mathcal D = \left\{\bm \Xi_{c,i}, \bm u_i\right\}_{i=1}^{N_r}$, where
    \begin{equation}
        \bm \Xi_x = \bm \xi_{1:N_s} \otimes \bm x_{1:N_x} \otimes \bm t_{1:N_t},
    \end{equation}
    and 
    \begin{equation}
        N_r = N_s\times N_x\times N_t.
    \end{equation}
    \item Represent the output $y$ using DNN,
    \begin{equation}
        u = \mathbb N\left(\bm \xi, x, t; \bm \theta \right).
    \end{equation}

    \item Compute the DNN parameters $\bm \theta$ by minimizing the loss-function in \autoref{eq:loss},
    \begin{equation}
        \bm \theta^* = \arg \min_{\bm \theta} \mathcal L_d \left( \bm \theta \right).
    \end{equation}
\end{itemize}
In PI-DNN, the objective is to remove the data-generation step and compute the DNN parameters
$\bm \theta$ directly from the governing differential equation in \autoref{eq:lf_gde}.
Following the method presented in \cite{chakraborty2020simulation}, this is achieved in
four simple steps.
First, similar to the data-driven case, the response $u$ is represented by using a DNN,
\begin{equation}\label{eq:dnn}
    u \approx u_{NN} = \mathbb N\left(\bm \xi, x, t; \bm \theta \right).
\end{equation}
Second, the neural network outputs are modified 
so as to automatically satisfy the initial and Dirichlet boundary conditions.
\begin{equation}\label{eq:bs_dnn}
    \hat u (\bm \xi, x, t) = u_b(x_b, t_i) + B\cdot u_{NN} (x,t,\bm \xi),
\end{equation}
where the function $B$ is defined in such a way that $B = 0$ at the boundary $(x_b)$ and initial $(t_i)$ points.
The function $u_b(x_b, t_i)$ is defined based on the initial and boundary conditions.
More details on this can be found in \cite{goswami2020transfer,chakraborty2020simulation}.
Note that $\hat u (\bm \xi, x, t)$ can also be viewed
as a DNN, $\hat{\mathbb N}(\bm \xi, x, t; \bm \theta)$.

In the third step, collocation points for the inputs, $\mathcal D_c = \left\{ \bm \xi_k, x_k, t_k \right\}_{k=1}^{N_c}$
are generated by using some suitable design of experiment scheme \cite{Chakraborty2016sequential,Bhattacharyya2018a}.
Using the collocation points, the physics-informed
loss function is formulated as
\begin{equation}\label{eq:pi_loss}
    \mathcal L_p(\bm \theta) = \frac{1}{N_c}\sum_{i=1}^{N_c}R_i^2, 
\end{equation}
where $N_c$ is the number of collocation points and
$R_i$ is the residual of the governing differential
equation corresponding to the $i-$th collocation point,
\begin{equation}\label{eq:residual}
    R_i = (\hat u_t)_i + h\left((\hat u)_i, (\hat u_x)_i, (\hat u_{xx})_i,\ldots; \bm \xi_i \right).
\end{equation}
$(\hat u)_i$ in \autoref{eq:residual} is obtained by substituting the $i-$th collocation point into \autoref{eq:bs_dnn}.
$(\hat u_t)_i$, $(\hat u_x)_i$, $(\hat u_{xx})_i$ are obtained by using automatic differentiation (AD) \cite{baydin2017automatic},
\begin{equation}\label{eq:dnn_derivatives}
    \begin{split}
        \hat u_t & = \frac{\partial \hat u}{\partial t} = \hat{\mathbb N}^t(\bm \xi, x, t; \bm \theta),\\
        \hat u_x & = \frac{\partial \hat u}{\partial x} = \hat{\mathbb N}^x(\bm \xi, x, t; \bm \theta),\\
        \hat u_{xx} & = \frac{\partial ^2 \hat u}{\partial x^2} = \hat{\mathbb N}^{xx}(\bm \xi, x, t; \bm \theta).
    \end{split}
\end{equation}
Note that the derivatives in \autoref{eq:dnn_derivatives} are also DNN.
Since the DNNs in \autoref{eq:dnn_derivatives} are 
obtained by differentiating \autoref{eq:bs_dnn},
they have the same architecture and same parameters;
the only difference is in the form of the activation function.

In the fourth and final step, the loss function in \autoref{eq:pi_loss} is minimized to compute the 
parameters of the DNN,
\begin{equation}\label{eq:train_pinn}
    \bm \theta^* = \arg \min_{\bm \theta} \mathcal L_p (\bm \theta).
\end{equation}
For further details on PI-DNN and its application in
solving reliability analysis problems, interested 
readers may refer
\cite{chakraborty2020simulation}.

PI-DNN has two major advantages.
First, unlike other reliability analysis tools including data-driven DNN, PI-DNN needs no simulation data. This is expected to reduce the computational cost significantly.
Second, PI-DNN is trained by satisfying the governing differential equation of the system.
Therefore, physical properties such as invariance and symmetries are satisfied.
However, despite these advantages, the whole idea of 
PI-DNN is hinged on the fact that the exact governing differential equation for 
the system under consideration is available.
Unfortunately, this is not necessarily true.
There exists a number of scenarios in science and 
engineering where the governing differential equation
is not known \cite{brunton2016discovering}.
Even if the governing equation is known, it is often
based on certain assumptions and approximations.
In other words, the governing differential equation
only represents the reality in an approximate manner.
Under such circumstances, results obtained using 
PI-DNN are bound to be erroneous.
\subsection{Proposed approach}
\label{subsec:pa}
Neither the data-driven DNN in \autoref{subsec:ddnn} nor the PI-DNN presented in \autoref{subsec:pinn}
is capable of solving the reliability analysis
problem defined in \autoref{sec:ps}.
The data-driven DNN fails because the number of high-fidelity data available, $N_h$ is very less.
On the other hand, the PI-DNN fails as the governing differential equation in \autoref{eq:lf_gde} only represents the actual scenario in an approximate manner.
To solve the problem defined in \autoref{sec:pa},
a multi-fidelity physics informed
deep neural network (MF-PIDNN) is presented in this section.
MF-PIDNN utilizes the concepts of both data-driven and physics informed DNNs.
Unlike available multi-fidelity frameworks, the proposed MF-PIDNN does not assume that generating low-fidelity data is trivial.
In fact, no low-fidelity data is needed for the MF-PIDNN presented here.

The key consideration of any multi-fidelity framework
is associated with discovering and exploiting the
relation between the low-fidelity and high-fidelity
model/data.
In most of the frameworks available in the literature,
this is achieved by using two surrogates; the first surrogate is trained based on the low-fidelity data
and the second surrogate is used to find the functional relation between the low-fidelity and the high-fidelity data.
This paper takes a separate route;
instead of using two DNNs, a single DNN is first trained for the low-fidelity model and then updated
based on the high-fidelity data.
For updating the DNN, the concept of transfer learning is used in this study.
Note that the idea of using transfer learning 
in a multi-fidelity framework has previously been 
exploited in \cite{de2020transfer}.
However, unlike the proposed framework, the algorithm
presented in \cite{de2020transfer} is purely data-driven in nature.

\begin{figure}[ht!]
    \centering
    \subfigure[Low-fidelity training phase]{
    \includegraphics[width = 0.8\textwidth]{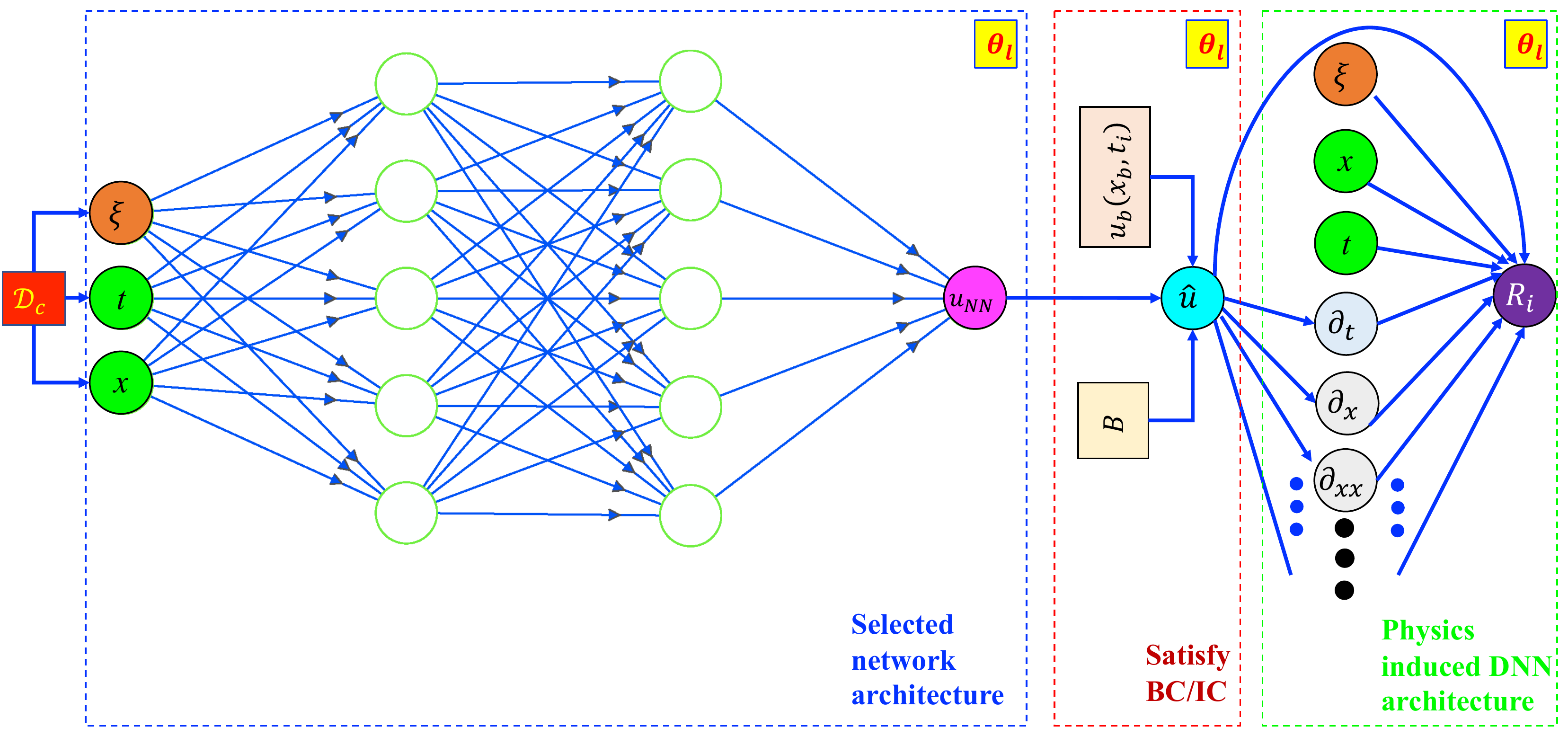}}
    \subfigure[High-fidelity training phase]{\includegraphics[width = 0.6\textwidth]{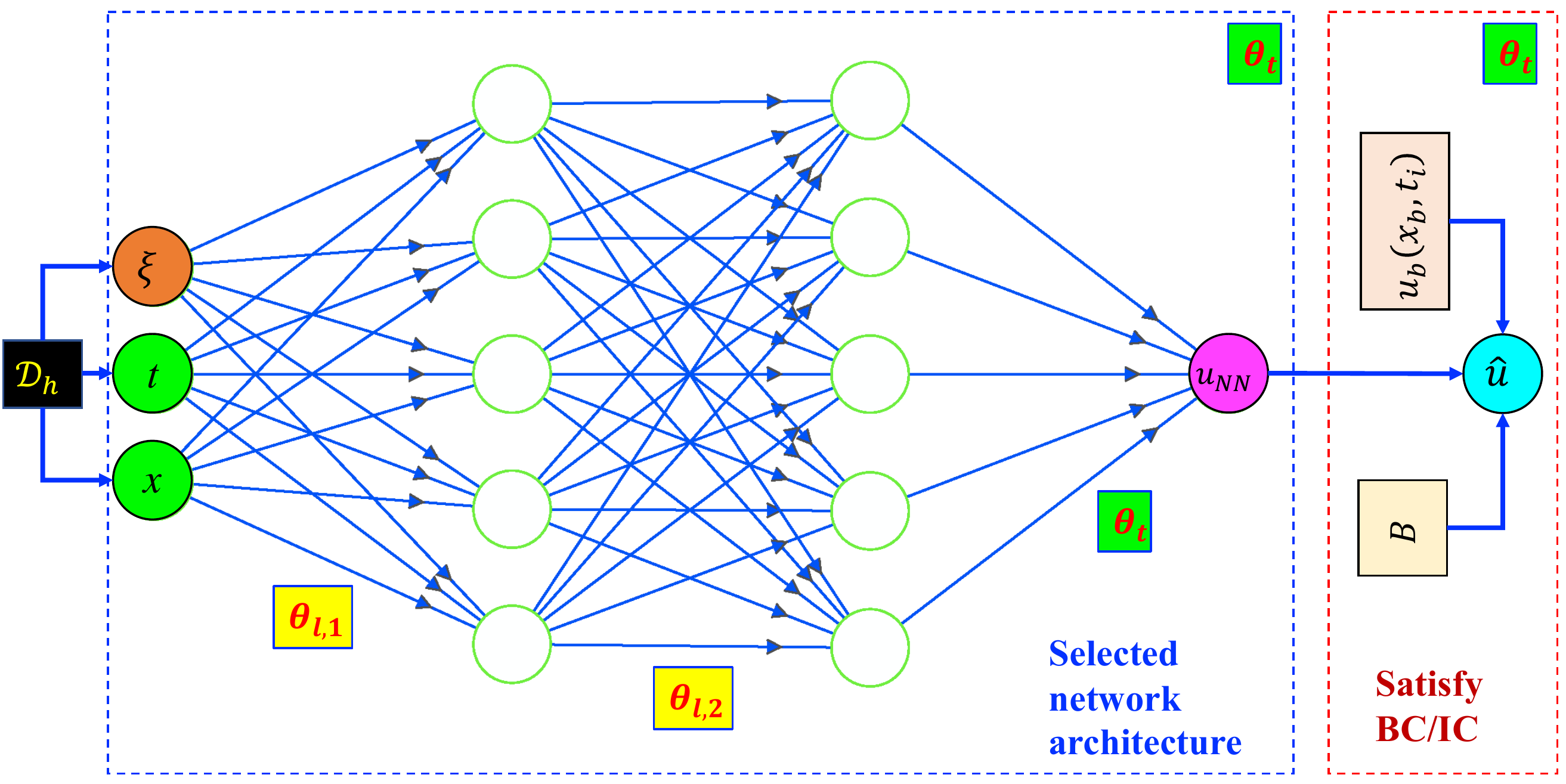}}
    \caption{Schematic representation of the proposed MF-PIDNN. During the low-fidelity training phase in (a), the DNN has three building blocks. The physics induced DNN architecture is governed by the low-fidelity governing differential equation of the system. At this stage, the DNN parameters are tuned by using the collocation points $\mathcal D_c $ and minimizing the residual, $(R_i)$ (physics-informed loss). $\bm \theta_l$ (yellow box) indicates that the DNN parameters obtained at the end of the training phase. During the high-fidelity training phase in (b), the DNN parameters for all but the last one or two layers are fixed at $\bm \theta_l$ (yellow box)). The tunable parameters $\bm \theta_t$ (green box) are estimated by minimizing the mean-squared error computed using the high-fidelity data $\mathcal D_h$.}
    \label{fig:schematic}
\end{figure}

MF-PIDNN solves the problem defined in \autoref{sec:ps} in two simple steps.
In the first step, PI-DNN is used to solve the
low-fidelity model.
To that end, the exact procedure as discussed in \autoref{subsec:pinn} is followed.
In the second step, the low-fidelity PI-DNN is 
updated based on the high-fidelity data $\mathcal D_{hx}$.
This is achieved by utilizing the concept of data-driven DNN.
However, unlike the first step, the second step is not straight-forward.
More specifically, two specific factors are considered
in this step.
First, the training algorithm starts by setting the 
initial value of the neural network parameters
to those obtained in step 1.
Second, the parameters corresponding to all the layers
are not updated.
Instead, the concept of transfer learning \cite{goswami2020transfer} is used and the parameters
corresponding to only the last one or two layers are updated.
A schematic representation of MF-PIDNN
is shown in \autoref{fig:schematic}.
The advantage of transfer learning is three-folds.
\begin{itemize}
    \item First, because of transfer learning, the number of
parameters to be updated is reduced. This in turn, 
accelerates training of the DNN.

\item Second, freezing the parameters of the initial layer ensures that the features learned/extracted from the 
low-fidelity model are retained in the network.

\item Thirdly, transfer learning also ensures that
the DNN does not overfit the high-fidelity data, $\mathcal D_h$. 
\end{itemize}
The steps involved in the proposed MF-PIDNN are
shown in \autoref{alg:mfpidnn}.
\begin{algorithm}[ht!]
 \caption{Transfer learning based multi-fidelity physics informed deep neural network}\label{alg:mfpidnn}
 \textbf{Initialize: }Provide high-fidelity data $\mathcal D_h$ and the low-fidelity model. Also specify the architecture of the DNN and the number
 of tunable layers, $l_t$ during transfer learning. \\
 Express the unknown response using a DNN \Comment*[r]{\autoref{eq:dnn}} 
 Modify the DNN to automatically satisfy the initial
 and boundary conditions \Comment*[r]{\autoref{eq:bs_dnn}} 
 Utilize the low-fidelity physics to formulate a physics-informed loss function \Comment*[r]{\autoref{eq:pi_loss}}
 Minimize the physics-informed loss function to 
 compute the network parameters, $\bm \theta$
 \Comment*[r]{\autoref{eq:train_pinn}}
 Freeze the DNN parameters for initial $(L - l_t + 1)$ layers. \\
 Formulate data-driven loss function using $\mathcal D_h$ \Comment*[r]{\autoref{eq:loss}}
 Minimize the loss-function to tune the tunable parameters
 \[ \bm \theta_{t} = \arg \min_{\bm \theta_t} \mathcal L_d(\bm \theta_t),\]
where $\bm \theta_t$ represents the tunable parameters. 
\end{algorithm}
 For training the MF-PIDNN, RMSProp optimizer \cite{tieleman2012lecture} followed by L-BFGS algorithm is used.
 Xavier initialization is used for initializing the DNN parameters.
 Details on the parameters settings for the optimizers are provided in \autoref{sec:ni}.
 Once the MF-PIDNN is trained, it is possible to predict $u$ corresponding to some unknown inputs by using \autoref{eq:bs_dnn}.
\section{Numerical illustration}\label{sec:ni}
In this section, four numerical examples are presented to illustrate the 
performance of the proposed approach.
A wide variety of examples involving single and
multiple stochastic variables, linear and non-linear
problems, ordinary and partial differential equations
are selected.
For illustrating the performance of the proposed approach, benchmark results using Monte Carlo simulation (MCS) \cite{Rubinstein1981simulation} are generated.
The software accompanying the proposed approach is
developed using \texttt{TenserFlow} \cite{abadi2016tensorflow}.
For examples 1, 2 and 4, the benchmark results are generated using \texttt{MATLAB} \cite{matlab2019}.
Benchmark results for example 2 are generated using
the \texttt{FeNICS} package \cite{alnaes2015fenics}.
\subsection{An ordinary differential equation}\label{eg:ode}
As the first example, a benchmark stochastic ordinary differential equation previously studied in \cite{li2010evaluation} is considered.
The low-fidelity model for this problem is given by
the following stochastic ordinary differential equation,
\begin{equation}\label{eq:eg1_lf}
    \frac{\text d u_l}{\text d t} = - Z u_l,
\end{equation}
where $Z$ is the stochastic variable.
The differential equation in \autoref{eq:eg1_lf} is subjected to the following initial condition,
\begin{equation}\label{eq:eg1_bc}
    u_l\left(t=0\right) = 1.0.
\end{equation}
The high-fidelity model, on the other hand, is represented as
\begin{equation}\label{eq:eg1_hf}
    u_h = t\sin\left(t\right) \left[\log \left( u_l ^4 \right)\right]^2 + 15 t^3 + 1.0
\end{equation}
Clearly, the relation between the high-fidelity $u_h$
and the low-fidelity $u_l$ is non-linear.
The limit-state function for this problem is defined 
as
\begin{equation}\label{eq:eg1_ls}
    \mathcal J(Z,t_t) = u_h(Z,t_t) - u_0, 
\end{equation}
where $u_0$ is the threshold, $u_h(Z,t_t)$ is the
response and $t_t$ is the time at which the 
probability of failure is to be estimated.
For this example, $t_t = 1.0$ and $u_0 = 18.0$ is considered.
It is assumed that 15 samples from the high-fidelity model is available, and
for each of the 15 high-fidelity samples, the observations are available at $t=[0.0,1.0]$.
Note that the data-generation process, i.e., \autoref{eq:eg1_hf} is not known. MF-PIDNN only have 
access to the high-fidelity data and the low-fidelity model in \autoref{eq:eg1_lf}.
For this particular problem, the stochastic variable $Z \sim \mathcal{N}(\mu, \sigma^2)$ is considered to 
follow Gaussian distribution with mean $\mu = -2.0$
and standard deviation $\sigma = 1.0$.
MCS with $10^6$ simulations yields a probability of failure of 0.045.

For solving the problem using the proposed approach,
the unknown response $u$ is first represented by using an FC-DNN with 2 inputs, 5 hidden layers and 50 neurons
per hidden layer.
The 2 inputs to the DNN are time $t$ and decay parameter $Z$.
Hyperbolic tangent (\texttt{tanh}) activation function is considered for all but the last layer.
For the last layer, linear activation function is
considered.
The initial conditions in \autoref{eq:eg1_bc} is automatically satisfied by modifying the DNN output, $u_{NN}$ using \autoref{eq:bs_dnn}, where $u_b = 1.0$
and $B = t$,
\begin{equation}\label{eq:eg1_bsdnn}
    \hat u =  t \cdot u_NN + 1.0.
\end{equation}
The residual for training the low-fidelity DNN
is
\begin{equation}
    R_i = \frac{\text d \hat u_i}{\text d t} + Z_i \hat u_i,
\end{equation}
where $R_i$ is the residual and $\hat u_i$ is obtained from \autoref{eq:eg1_bsdnn}.
`$i'$ in the suffix indicates the $i-$th collocation point.
For training the low-fidelity model, $8000$ collocation points is used and the RMSprop optimizer is run for $15,000$ iterations. A learning rate of 0.001 is used. The other parameters of RMSprop are kept at there default values.
The maximum allowable iterations for L-BFGS optimizer
is set to be $10,000$.

After training the physics-informed low-fidelity DNN,
the next step is to update the model based on the high-fidelity data by using the transfer learning.
The parameters corresponding to the last two layers
are only updated; parameters corresponding to all 
other layers are kept fixed.
The RMSprop optimizer is run for $10,000$ iterations and
maximum allowable iterations for the L-BFGS optimizer
is set to be $10,000$.
For RMSProp optimizer, a learning rate of 0.001 is used.

\autoref{tab:eg1_res} shows the results obtained using
MCS and MF-PIDNN. Along with the probability of failure $P_f$, the reliability index $\beta$ for this problem is also reported.
\begin{equation}\label{eq:beta}
    \beta =  \Phi^{-1}\left(1 - P_f\right),
\end{equation}
where $ \Phi\left(\cdot\right)$ represents cumulative distribution function of standard Gaussian distribution.
The results obtained using MF-PIDNN matches exactly
with the MCS results.
To show the utility of the proposed approach, results obtained using only the low-fidelity PI-DNN and the high-fidelity DNN are also presented.
Both low-fidelity PI-DNN and high-fidelity DNN are found to yield erroneous results.
\begin{table}[ht!]
    \centering
    \caption{Reliability analysis results for example 1.}
    \label{tab:eg1_res}
    \begin{tabular}{lccccc}
    \hline
        \textbf{Methods} & $P_f$ & $\beta$ & $N_h$ & $N_r$ & $\epsilon = \frac{\left| \beta_e - \beta \right|}{\beta_e}\times 100$  \\ \hline
        MCS & $0.045$ & $1.6954$ & $10^6$ & $10^6 \times 1001 $ & -- \\ \hline \hline 
        LF-PIDNN & 0.8133 & -0.8901 & 0 & 0 & 152.5\% \\
        HF-DNN & 0.0 & $\infty$ & 15 & $30(15\times2)$ & $\infty$ \\
        MF-PIDNN & 0.045 & 1.6954 & 15 & $30(15\times2)$ & 0.0\% \\ \hline
    \end{tabular}
\end{table}

To further illustrate the performance of the MF-PIDNN, two additional case studies are performed.
In the first case study, the performance of the MF-PIDNN in predicting 
future reliability is investigated.
To that end, it is assumed that for each of the 15
high-fidelity samples, observations are available $t = [0.0, 0.5, 0.9]$, and
the objective is to compute the reliability of the system at $t=1.0$.
The difficulty, in this case, arises from the fact that
this is an extrapolation problem as no observation is available at or beyond $t=1.0$.
The network architecture and other parameters of
MF-PIDNN are considered to be same as before; the only difference resides in the fact that the RMSProp
optimizer is run for $15,000$ iterations  (while updating
the network using transfer learning).
The results obtained are shown in \autoref{tab:eg1_res2}.
Compared to the results presented in \autoref{tab:eg1_res}, slight deterioration in the 
results have been observed; this is expected because this is an extrapolation problem.
Nonetheless, the results obtained are still significantly more accurate as compared to HF-DNN and
LF-PIDNN.
\begin{table}[ht!]
    \centering
    \caption{Reliability analysis results for example 1. The results presented illustrate the extrapolation capability of the MF-PIDNN.}
    \label{tab:eg1_res2}
    \begin{tabular}{lccccc}
    \hline
    \textbf{Methods} & $P_f$ & $\beta$ & $N_h$ & $N_r$ & $\epsilon = \frac{\left| \beta_e - \beta \right|}{\beta_e}\times 100$ \\ \hline
       MCS & $0.045$ & $1.6954$ & $10^6$ & $10^6 \times 1001$ & -- \\ \hline \hline 
        LF-PIDNN & 0.8133 & -0.8901 & 0 & 0 & 152.5\% \\
        HF-DNN & 0.014 & 2.9173 & 15 & $45(15\times3)$ & 29.60\% \\
        MF-PIDNN & 0.05 & 1.6449 & 15 & $45(15\times3)$ & 2.98\% \\ \hline
    \end{tabular}
\end{table}

\begin{figure}[ht!]
    \centering
    \includegraphics[width = 0.6\textwidth]{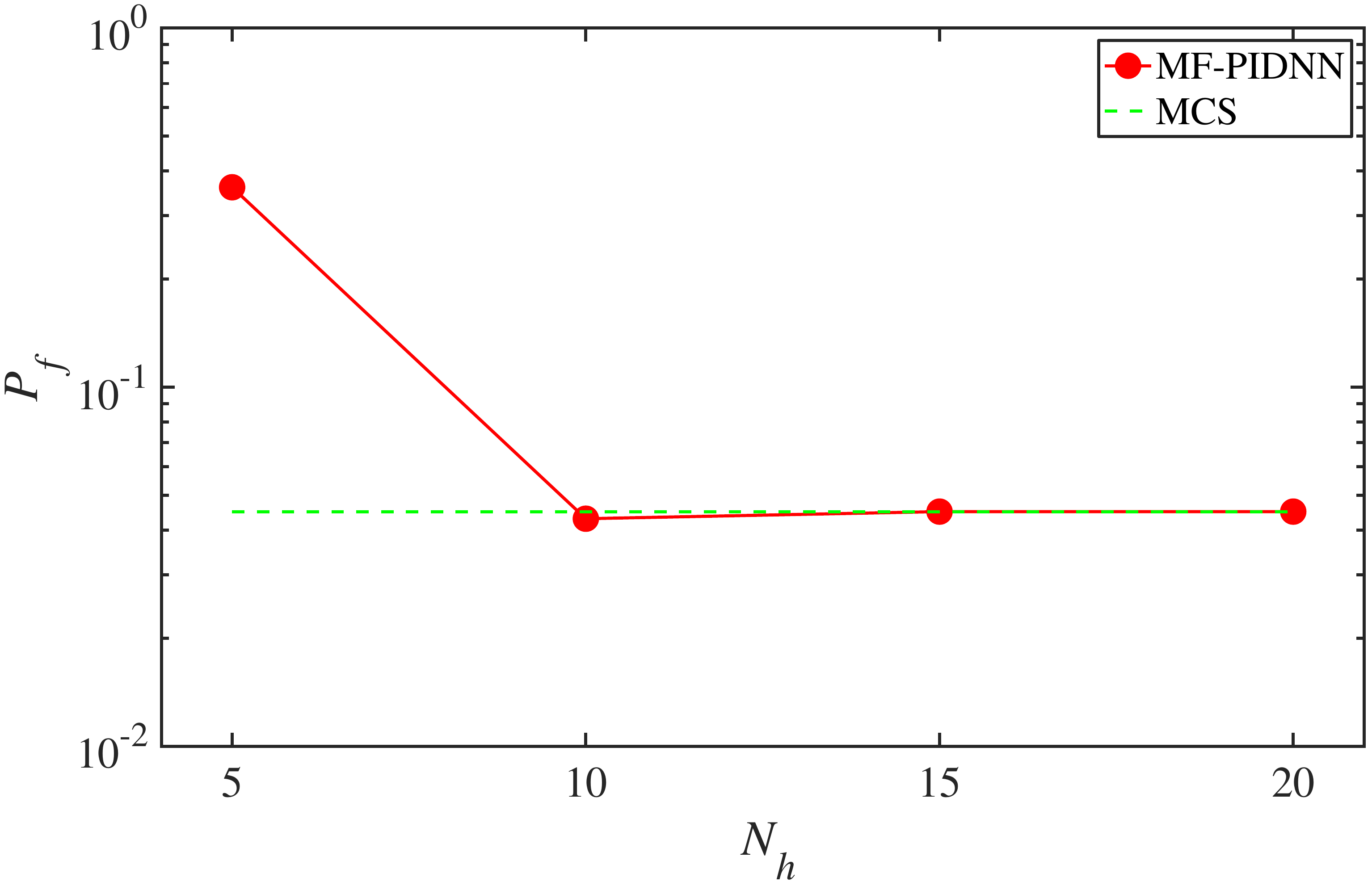}
    \caption{Variation in the MF-PIDNN predicted results with increase in number of high-fidelity data points.}
    \label{fig:eg1}
\end{figure}

Finally, the performance of the MF-PIDNN with variation in the number of high-fidelity data point, $N_h$ is investigated. 
For each realization of $Z$, the responses are observed at $t=[0.0,1.0]$ and the probability of failure at $t_t= 1.0$ is computed.
The variation of the MF-PIDNN predicted probability of failure is shown in \autoref{fig:eg1}.
The benchmark result obtained using MCS is also reported.
With an increase in $N_h$, the MF-PIDNN predicted probability of failure converges to the MCS solution.
The HF-DNN results, up to $N_h = 20$, yields erroneous results (not shown in \autoref{fig:eg1}). This is because, with only observations at
two time-instants, the DNN fails to predict the trend
of the limit-state function.
MF-PIDNN, on the other hand, learns the trend from the physics of the problem and then update itself 
based on the high-fidelity data.
\subsection{Burger's equation}
As the second problem, the well-known Burger's equation is considered. 
The high-fidelity model for this problem is
\begin{equation}\label{eq:eg2_hf}
    (u_h)_t + u_h (u_h)_x = \nu (u_h)_xx,
\end{equation}
with $x \in \left[ -1, 1 \right]$ and $t \in \left[0,12\right]$.
$\nu > 0$ in \autoref{eq:eg2_hf} represents the viscosity.
The boundary and the initial conditions for this problem are
\begin{subequations}\label{eq:eg2_bc_and_ic}
\begin{equation}\label{eq:eg2_bc}
    u_h(t,x = -1) = 1 + \delta\;\;\;\;u_h(t,x=1) = -1,
\end{equation}
\begin{equation}\label{eq:eg2_ic}
    u_h(t=0,x) = -1 + (1 + x )\left( 1 + \frac{\delta}{2} \right).
\end{equation}
\end{subequations}
$\delta$ in \autoref{eq:eg2_bc_and_ic} is a small perturbation that is applied to the left boundary.
The problem as defined has a transition layer at $z$,
so that $u_h(z) = 0$.
As illustrated in previous studies \cite{xiu2004supersensitivity,lorenz1981nonlinear}, the transition
layer is super sensitive to $\delta$.
Details on different aspects of this problem can be found in \cite{xiu2004supersensitivity,lorenz1981nonlinear}.

The low-fidelity model, on the other hand, is considered to be
\begin{equation}\label{eq:eg2_lf}
    (u_l)_t = \nu (u_l)_{xx}.
\end{equation}
\autoref{eq:eg2_lf} is obtained by ignoring the nonlinear term in the high-fidelity model.
The initial and the boundary conditions are considered to be same as the high-fidelity model.
The boundary perturbation $\delta \sim \mathcal U \left(0.0, 0.1\right)$ is uniformly 
distributed between 0.0 and 0.1.
It is considered that the high-fidelity model
in \autoref{eq:eg1_hf} is not known; instead,
data corresponding to five realizations of $\delta$
is available.
For each $\delta$, observations at 3 spatial location
and 8 temporal locations are available.
\begin{equation}\label{eq:eg2_hfdata}
    \bm \Xi_{hx} = \left[0,0.025,0.05,0.075,0.1 \right] \otimes \left[-1, 0, 1 \right] \otimes \left[1, 2.14, 3.29, 4.43, 5.57, 6.71, 7.86, 9 \right].
\end{equation}

The limit-state function for this problem is represented as
\begin{equation}\label{eq:eg2_ls}
    J\left(\delta,t_t\right) = -z(\delta, t) + z_0,
\end{equation}
where $z$ represents the transition layer, $t_t$ is the time at which the reliability is to be computed and $z_0$ is the threshold. For this example, $z_0 = 0.40$ is considered.
The objective is to compute the probability of failure at $t_t=10$.
Note that solution of this problem involves extrapolation as no observation at $t = 10$ or
beyond is available.

For solving the problem using the proposed MF-PIDNN,
$u$ is first represented by using a FC-DNN with $6$ hidden layers.
Each of the 6 hidden layers has 50 neurons.
The DNN has 3 inputs, $x$, $t$ and $\delta$ and one 
output $u_{NN}$.
\texttt{tanh} 
activation function is considered for all but the last layer. 
For the last layer, linear activation function is used.
To automatically satisfy the boundary and initial 
conditions, the DNN output is modified as
\begin{equation}
    \hat u = u_h(t=0,x) + t(1-x)(1+x)u_{NN},
\end{equation}
where $u_h(t=0,x)$ is obtained from \autoref{eq:eg2_ic}.
Using $\hat u$ and its derivatives, the residual
of the low-fidelity model is formulated as
\begin{equation}\label{eq:eg2_residual}
    R_i = (\hat u_t)_i - \nu (\hat u)_{xx})_i,
\end{equation}
where $R_i$ is the residual.
$i$ in the suffix indicates that the quantities
are evaluated corresponding to the $i-$th collocation point.
The physics-informed loss-function for training the 
low-fidelity model is formulated by using $30,000$
collocation points and \autoref{eq:eg2_residual}.
Because of the simplicity of the low-fidelity model,
the RMSProp optimizer is run for 500 iterations and the maximum allowable iterations for the L-BFGS
optimizer is set to be 1000. The learning rate in RMSProp optimizer is set to be 0.001.
Once the low-fidelity physics informed DNN is trained,
the next step is to update the DNN model by using
transfer learning.
To retain information gained from the low-fidelity model and avoid over-fitting, parameters corresponding to only the 
last two layers of the DNN are allowed to update; all
the other parameters are frozen.
The RMSprop optimizer is run for 6000 iterations with a learning rate of 0.003.
The maximum allowed iterations for the L-BFGS algorithm is set to 10,000.
The L-BFGS optimizer is only allowed to update the
DNN parameters corresponding to the last layer.
For this problem, the MF-PIDNN is found to be highly sensitive to the initial point of the parameters and
varies from run to run.
Therefore, the MF-PIDNN results presented
are mean predictions after running the model
for 20 times.

For the purpose of validation, benchmark results 
using MCS with $10^4$ simulations are generated.
To that end, finite element package \texttt{FeNICS} \cite{alnaes2015fenics} is used. The same-solver is
used for generating the high-fidelity data as well.

\begin{figure}[ht!]
    \centering
    \includegraphics[width = 0.6\textwidth]{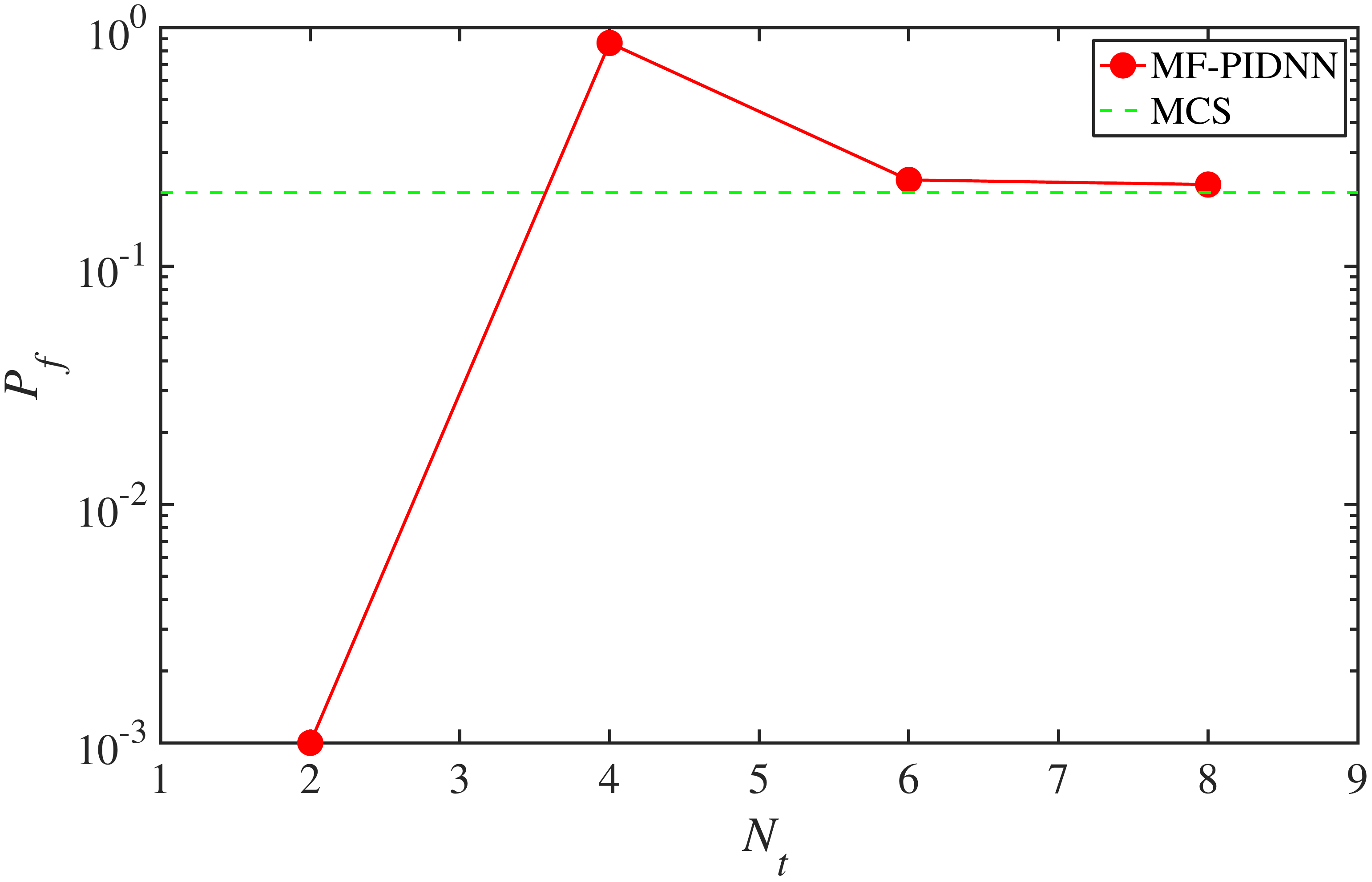}
    \caption{Variation of MF-PIDNN predicted results with $N_t$.}
    \label{fig:eg2_conv}
\end{figure}

The reliability analysis results are shown in \autoref{tab:eg2_res}. Along with MCS and MF-PIDNN results,
LF-PIDNN and HF-DNN predicted results are
also presented. 
Similar to the previous example, both probability of
failure and reliability index are reported.
It is observed that MF-PIDNN predicted results are extremely close to the MCS results. HF-DNN and LF-PIDNN, on the other hand, yields erroneous results. 
\autoref{fig:eg2_conv} shows the performance of MF-PIDNN with increase in $N_t$ (i.e, number of time-steps at which high-fidelity data is available).
It is observed that with an increase in $N_t$, the 
MF-PIDNN predicted result moves closer to the MCS results. However, at $N_t = 6$ and 8, the probability of failure obtained is found to be similar, indicating convergence of the proposed approach.
\begin{table}[ht!]
    \centering
    \caption{Reliability analysis results for the Burger's equation}
    \label{tab:eg2_res}
    \begin{tabular}{lccccc}
    \hline 
    \textbf{Methods} & $P_f$ & $\beta$ & $N_h$ & $N_r$ & $\epsilon = \frac{\left| \beta - \beta_e \right|}{\beta_e}\times 100$ \\ \hline
    MCS & 0.2036 & 0.8288 & $10^4$ & $10^4 \times 33 \times 10^3$ & --  \\ \hline \hline
    LF-PIDNN & 0 & $\infty$ & 0 & 0 & $\infty$ \\
    HF-DNN & 0.932 & -1.4909 & 5 & $120 (5 \times 3 \times 8)$ & 280\% \\
    MF-PIDNN & 0.2242 & 0.7581 & 5 & $120 (5 \times 3 \times 8)$ & 8.5304\% \\
    \hline
    \end{tabular}
\end{table}

\subsection{Nonlinear oscillator}
\label{subsec:eg3}
As the third example, a nonlinear oscillator, previously studied in \cite{qin2019neural} has been considered.
The high-fidelity model for this problem is given as
\begin{equation}\label{eq:eg3_hf}
    \begin{split}
        \frac{\text d (x_h)_1}{\text d t} & = (x_h)_2, \\
        \frac{\text d (x_h)_2}{\text d t} & = -\alpha_1 (x_h)_2 - \alpha_2 \sin \left((x_h)_1\right),
    \end{split}
\end{equation}
where $\alpha_1$ and $\alpha_2$ are the stochastic parameter.
The initial conditions for the problem are
\begin{equation}\label{eq:eg3_ic}
    (x_h)_1(t = 0) = -1.193,\;\;\; (x_h)_2(t=0) = -3.876.
\end{equation}
The low-fidelity model, on the other hand, is given as
\begin{equation}\label{eq:eg3_lf}
\begin{split}
    \frac{\text d (x_l)_1}{\text d t} & = (x_l)_2, \\
        \frac{\text d (x_l)_2}{\text d t} & = -\alpha_1 (x_l)_2 - \alpha_2  (x_h)_1.
        \end{split}
\end{equation}
The initial condition for the low-fidelity model is considered to be same as the high-fidelity model.
Similar to the previous examples, 
the high-fidelity equation is assumed to be unknown and one 
only has access to the low-fidelity model and high-fidelity data.
More specifically, data corresponding to five realizations of the stochastic parameters
are available.
For each of the realizations, the observations are available at five equally spaced time-instants in $[0,5]$.
The realizations of the stochastic parameters are 
obtained using Latin hypercube sampling \cite{Iman1980latin}.
Following \cite{qin2019neural}, the stochastic parameters $\alpha_1 \sim \mathcal U \left( 0, 0.4 \right)$
and $\alpha_2 \sim \mathcal U \left( 8.8, 9.2 \right)$ are considered to be uniformly distributed.
The limit-state function for this problem is defined
as
\begin{equation}\label{eq:eg3_ls}
    \mathcal J (\alpha_1, \alpha_2, t_t) = - \left| x_2 (\alpha_1, \alpha_2, t_t)\right| + x_0,
\end{equation}
where $x_0$ is the threshold and $t_t$ is the time
at which the reliability is to be evaluated.
For this problem, $t_t = 5.0$ and $x_0 = 4.0$ is 
considered.

To solve the problem using MF-PIDNN, $x_i,\,i=1,2$ is first represented using a FC-DNN having 4 hidden 
layers. Each hidden layer has 50 neurons.
The DNN has three inputs, $t$, $\alpha_1$ and $\alpha_2$ and two outputs $x_1$ and $x_2$.
\texttt{tanh} 
activation function is considered for all but the last layer. 
Linear activation function is used for the last layer.
To automatically satisfy the initial conditions,
the DNN output is modified as
\begin{equation}
    \begin{split}
        \hat x_1 & = t\cdot x_{NN,1} -1.193,\\
        \hat x_2 & = t\cdot x_{NN,2} - 3.876,
    \end{split}
\end{equation}
where $x_{NN,1}$ and $x_{NN,2}$ are the DNN outputs.
Using $\hat x_1$ and $\hat x_2$, the residuals
are computed,
\begin{equation}\label{eq:eg3_residual}
    \begin{split}
        R_{1,i} & = \left((\hat x_1)_t\right)_i - (\hat x_2)_i,\\
        R_{2,i} & = \left((\hat x_2)_t\right)_i + (\alpha_1)_i (\hat x_2)_i + (\alpha_2)_i (\hat x_1)_i.
    \end{split}
\end{equation}
$i$ in \autoref{eq:eg3_residual} indicates the $i-$th collocation point.
Using the residuals, the physics-informed loss function for training the low-fidelity model is
formulated as
\begin{equation}
    \mathcal L_p (\bm \theta_l ) = \frac{1}{N_c}\sum_{i=1}^{N_c}\left(R_{1,i}^2 + R_{2,i}^2 \right),
\end{equation}
where $N_c$ is the number of collocation points.
For this problem, $10000$
collocation points have been used.
The RMSProp optimizer is run for 15000 iterations with a learning-rate of 0.001.
For L-BFGS optimizer, the maximum allowable iterations is set to 10000.
The trained low-fidelity model is then updated by using transfer learning and high-fidelity data.
Only the parameters corresponding to the last two layers of DNN are allowed to be updated.
A learning rate of 0.001 is used and the RMSProp 
optimizer is run for 10000 iterations. Maximum
allowable iterations for the L-BFGS optimizer 
is set to be 10000.

The benchmark results for validation are generated
by using MCS with $10^4$ simulations. To that end,
the differential equations are solved using the 
\texttt{ODE45} routine in \texttt{MATLAB} \cite{matlab2019}. The high-fidelity data-set discussed earlier was also generated by using the 
same solver.

\begin{figure}[ht!]
    \centering
    \includegraphics[width = 0.6\textwidth]{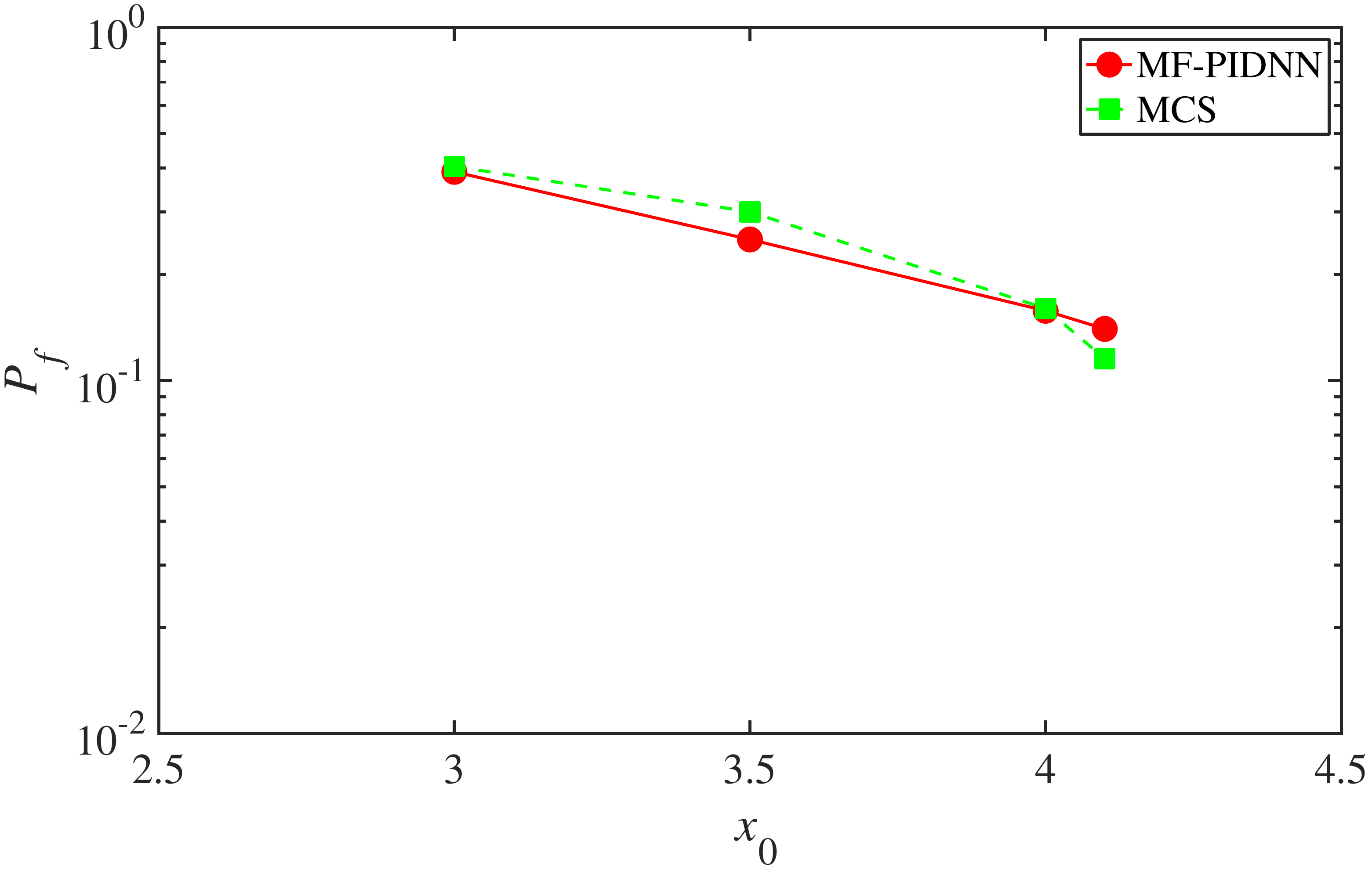}
    \caption{Variation of probability of failure with threshold $x_0$.}
    \label{fig:eg3_var}
\end{figure}

\autoref{tab:eg3_res} shows the reliability analysis
results for the nonlinear oscillator problem.
Similar to the previous examples, results obtained
using HF-DNN and LF-PIDNN are also presented.
The MF-PIDNN is found to yield highly accurate results, matching closely with the MCS solutions.
LF-PIDNN and HF-DNN yield erroneous results.
The variation of probability of failure with threshold $x_0$ is shown in \autoref{fig:eg3_var}.
Corresponding to all the thresholds, the MF-PIDNN
predicted results matches closely with the MCS
results. This indicates that the proposed MF-PIDNN is
able to capture the response over the whole domain.

\begin{table}[ht!]
    \centering
    \caption{Reliability analysis results for nonlinear oscillator.}
    \label{tab:eg3_res}
    \begin{tabular}{lccccc}
    \hline
         \textbf{Methods} & $P_f$ & $\beta$ & $N_h$ & $N_r$ & $\epsilon = \frac{\left| \beta - \beta_e \right|}{\beta_e}\times 100$ \\ \hline
    MCS & 0.1599 & 0.9949  & 10000 & $10^4 \times 10^3$  & --  \\ \hline \hline
    LF-PIDNN & 0.27  & 0.6128 & 0 & 0 & 38.41\% \\
    HF-DNN & 0.19 &  0.8779 & 5  & $5\times 5$ &  11.76\%  \\
    MF-PIDNN & 0.1576 & 1.0044  & 5  & $5\times 5$ & 0.95\% \\ \hline
    \end{tabular}
\end{table}

Lastly, to illustrate the robustness of the proposed MF-PIDNN, the same model is used to compute the
probability of failure at a $t_t = 3.0$ and $x_0 = 2.0$.
The results obtained are shown in \autoref{tab:eg3_res2}.
In this case also, MF-PIDNN is found to yield accurate results outperforming both HF-DNN and LF-PIDNN.

\begin{table}[ht!]
    \centering
    \caption{Reliability analysis results for nonlinear oscillator at $t_t = 3.0$ and $x_0 = 2.0$}
    \label{tab:eg3_res2}
    \begin{tabular}{lccccc}
    \hline
         \textbf{Methods} & $P_f$ & $\beta$ & $N_h$ & $N_r$ & $\epsilon = \frac{\left| \beta - \beta_e \right|}{\beta_e}\times 100$ \\ \hline
    MCS & 0.0651 & 1.5133  & 10000 & $10^4 \times 10^3$  & --  \\ \hline \hline
    LF-PIDNN & 0.98  & $-2.0537$ & 0 & 0 & 235.7\% \\
    HF-DNN & 0.5955 & $-0.2417$  &  5 & $5\times 5$ &  115.97\% \\
    MF-PIDNN & 0.0729 & 1.4545  & 5  & $5\times 5$ & 3.88\% \\ \hline
    \end{tabular}
\end{table}
\subsection{Cell signaling cascade}
\label{subsec:eg4}
As the last example, a mathematical model of autocrine cell-signaling cascade is considered.
\begin{equation}\label{eq:eg4_hf}
\begin{split}
    \frac{\text d e_{1_p}}{\text d t} & = \frac{I}{1 + G_4 e_{3_p}}\frac{V_{\max , 1}\left( 1 - e_{1_p}\right)}{K_{m,1} + (1 - e_{1_p})} - \frac{V_{\max , 2}e_{1_p}}{K_{m,2} + e_{1_p}},\\
    \frac{\text d e_{2_p}}{\text d t} & = \frac{V_{\max , 3} e_{1_p}\left( 1 - e_{2_p}\right)}{K_{m,3} + (1 - e_{2_p})} - \frac{V_{\max , 4}e_{2_p}}{K_{m,4} + e_{2_p}}, \\
    \frac{\text d e_{3_p}}{\text d t} & = \frac{V_{\max , 5} e_{2_p}\left( 1 - e_{3_p}\right)}{K_{m,5} + (1 - e_{3_p})} - \frac{V_{\max , 6}e_{3_p}}{K_{m,6} + e_{3_p}}, \;\;t \in [0,10],
\end{split}
\end{equation}
where $e_{1_p}$, $e_{2_p}$ and $e_{3_p}$ are the state variables and denotes concentrations of the active form of enzymes. $I$ in \autoref{eq:eg4_hf} is the tuning parameter.
The initial conditions for this problem are
\begin{equation}\label{eq:eg4_ic}
    e_{1_p}(t=0) = 0,\; e_{2_p}(t=0) = 1.0,\; e_{3_p}(t=0) = 0.
\end{equation}
This model was first developed in \cite{shvartsman2002autocrine}.
Overall the model has 13 parameters, $K_{m,1:6}$, $V_{\max , 1:6}$ and $G_4$.
For biological meaning and other details on the model parameters, interested readers may refer \cite{shvartsman2002autocrine}.

For reliability analysis, all the 13 parameters defined above are considered to be stochastic.
The mean of the parameters are adopted from \cite{shvartsman2002autocrine} and a 10\% relative noise is
added.
For clarity of readers, the mean of the 13 parameters is presented in \autoref{tab:eg4_mean_param}.
The same parameter settings have previously been used
in \cite{qin2019neural}.

\begin{table}[ht!]
    \centering
    \caption{Mean of the parameters for the cell signaling cascade problem}
    \label{tab:eg4_mean_param}
    \begin{tabular}{l|cccccccc}
    \hline
    \textbf{Parameters} & $K_{m,1:6}$ & $V_{\max ,1}$ & $V_{\max ,2}$ & $V_{\max ,3}$ & $V_{\max ,4}$ & $V_{\max ,5}$ & $V_{\max ,6}$ & $G_4$ \\ \hline
    \textbf{Mean} & 0.2 & 0.5 & 0.15 & 0.15 & 0.15 & 0.25 & 0.05 & 2 \\ \hline
    \end{tabular}
\end{table}

A low-fidelity model for this problem is set up by
considering $I = 0$.
With this, the coupled differential equations in
\autoref{eq:eg4_hf} is decoupled and it becomes 
possible to solve the equations sequentially.
Moreover, the stochastic variables $G_4$, $V_{\max ,1}$ and $K_{m,1}$ become inactive. This further
complicates the problem.
It is further assumed that the governing differential equation in \autoref{eq:eg4_hf} is not
available; instead, responses corresponding to 10
realizations of the stochastic variables are
available. 
For each of the 10 realizations, observations are
available at $5$ time-steps.
The observation time-instants are equally spaced
in $[4,7]$

The limit-state function for this problem is
\begin{equation}\label{eq:eg4_ls}
    \mathcal J \left( \bm \xi \right) = e_{3_p}(\bm \xi, t_t) - e_{3,0},
\end{equation}
where $\bm \xi \in \mathbb R^{13}$ represents the stochastic variables, $e_{3,0}$ is the threshold
parameter and $t_t$ is the time-instants at which 
the reliability is to be estimated.
For this problem, $t_t = 3.0$ is considered.
Since all the high-fidelity observation are available in $[4,7]$, this is an extrapolation problem.

For reliability analysis using MF-PIDNN, the output responses are first represented by using a FC-DNN.
The DNN has 14 inputs (13 stochastic variables and time), 3 outputs and 4 hidden layers.
Each of the hidden layers has 100 neurons.
All but the last layer of the DNN have \texttt{tanh}
activation function. For the last layer, 
linear activation function is used.
To automatically satisfy the initial conditions,
the DNN outputs are modified as
\begin{equation}\label{eq:eg4_bsdnn}
    \begin{split}
        \hat e_{1_p} & = t\cdot e_{1_p,NN} \\
        \hat e_{2_p} & = t\cdot e_{2_p,NN} + 1.0, \\
        \hat e_{3_p} & = t\cdot e_{3_p,NN},
    \end{split}
\end{equation}
where $e_{1_p,NN}$, $e_{2_p,NN}$ and $e_{3_p,NN}$
are the DNN outputs.
The residuals for formulating the physics-informed loss function are given as
\begin{equation}\label{eq:eg4_residuals}
\begin{split}
    R_{1,i} & = \left((K_{m,2})_i + (\hat e_{1_p})_i \right)((\hat e_{1_p})_t)_i + (V_{\max , 2})_i(\hat e_{1_p})_i, \\
    R_{2,i} & = \left((K_{m,4})_i + (\hat e_{2_p})_i \right)((K_{m,3})_i + (1 - (\hat e_{2_p})_i)) - (V_{\max , 3})_i (\hat e_{1_p})_i (1 - (\hat e_{2_p})_i), \\
    R_{3,i} & = \left((K_{m,6})_i + (\hat e_{3_p})_i \right)((K_{m,5})_i + (1 - (\hat e_{3_p})_i)) - (V_{\max , 5})_i (\hat e_{2_p})_i (1 - (\hat e_{3_p})_i),
\end{split}
\end{equation}
where `$i$' in suffix represents the $i-$th collocation point.
The residuals in \autoref{eq:eg4_residuals} corresponds to the low-fidelity model and hence, $I$, $G_4$ and $V_{\max ,1}$ are not present.
Using the residuals, the physics-informed loss
function for the low-fidelity model is computed
as
\begin{equation}\label{eq:eg4_pi_loss}
    \mathcal L_p (\bm \theta_l ) = \frac{1}{N_c}\sum_{i=1}^{N_c}\sum_{k=1}^3 R_{k,i}^2,
\end{equation}
where $N_c$ represents the number of collocation points.
For minimizing $\mathcal L_p(\bm \theta)$, the RMSProp optimizer is run for 5000 iterations.
A learning rate of 0.001 is used.
As for the L-BFGS optimizer, the maximum allowed 
iterations is set to 10000.
The trained low-fidelity model is then updated by using the high-fidelity data and transfer learning.
At this stage, only the parameters corresponding to 
the last layer is allowed to be tuned. 
All the other
parameters are fixed at $\bm \theta_l$.
A learning rate of 0.001 is used
and the RMSProp optimizer is run for 5000 iterations.
As for the L-BFGS optimizer, the maximum allowed iterations is set to be 10000.

The benchmark results for this problem are generated
by using MCS with $10^4$ simulations.
To that end, the \texttt{ODE45} routine available 
in \texttt{MATLAB} is used.
The high-fidelity data discussed before were also generated by using the same procedure.

\begin{figure}[ht!]
    \centering
    \includegraphics[width = 0.6\textwidth]{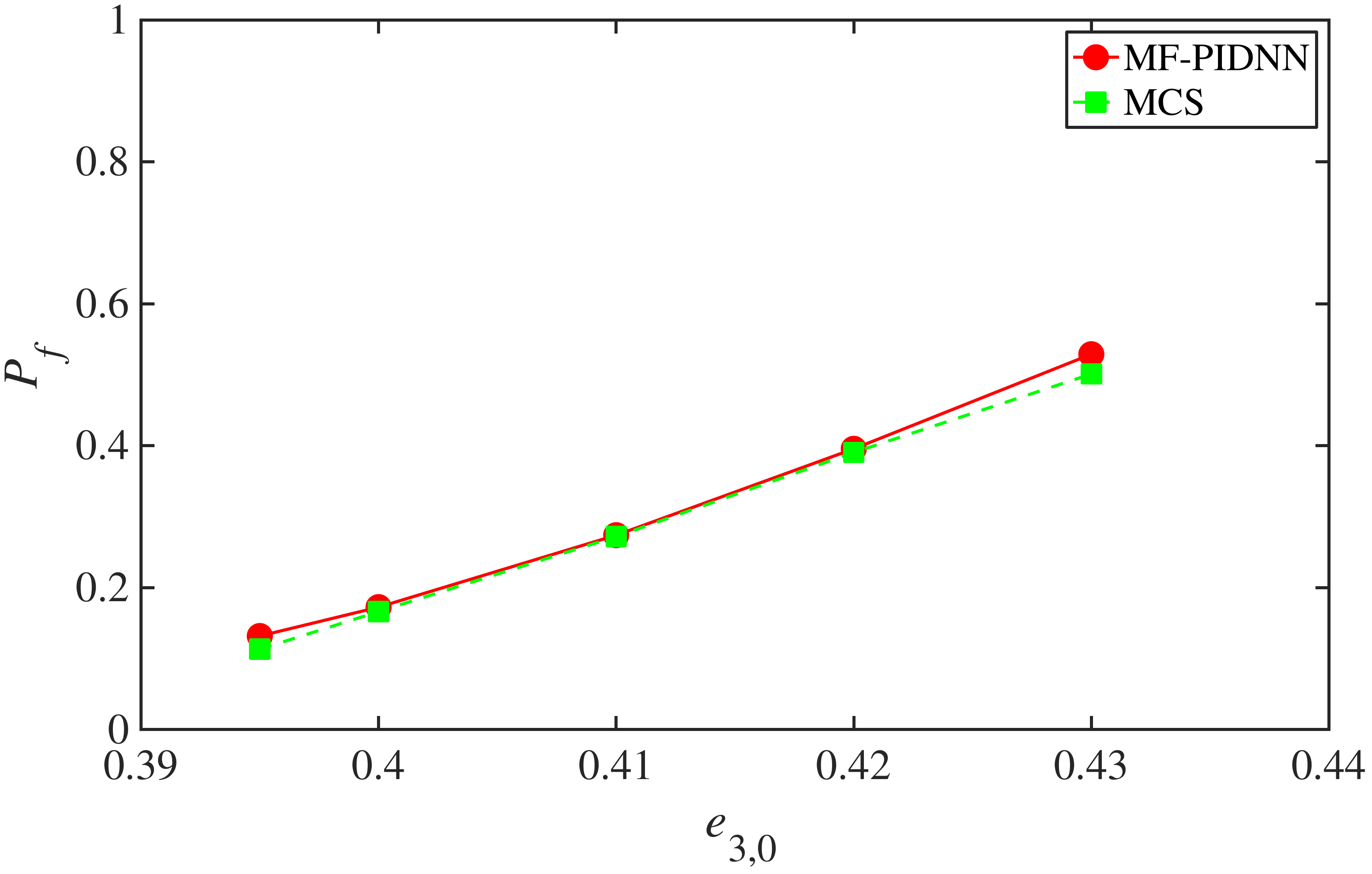}
    \caption{Variation of the probability of failure with threshold $e_{3,0}$.}
    \label{fig:eg4_conv}
\end{figure}

\autoref{tab:eg4_res} shows the reliability analysis
results for the cell signaling cascade problem.
Along with MCS and MF-PIDNN, results obtained using
LF-PIDNN and HF-DNN are also presented.
The proposed MF-PIDNN is found to yield highly accurate results with a prediction error of 1.52\%.
Results obtained using LF-PIDNN and HF-DNN respectively have an error of 64.35\% and 98.04\%.
The variation of probability of failure with the change
in threshold $e_{3,0}$ is shown in \autoref{fig:eg4_conv}. 
For all the thresholds, MF-PIDNN predicted results
are found to closely match with the MCS results.
This indicates that MF-PIDNN is able to capture the
response over the whole domain.

\begin{table}[ht!]
    \centering
    \caption{Reliability analysis results for cell signaling cascade problem}
    \label{tab:eg4_res}
    \begin{tabular}{lccccc}
    \hline
         \textbf{Methods} & $P_f$ & $\beta$ & $N_h$ & $N_r$ & $\epsilon = \frac{\left| \beta - \beta_e \right|}{\beta_e}\times 100$ \\ \hline
    MCS & 0.1663 & 0.9689  & 10000 & $10^4 \times 10^3$  & --  \\ \hline \hline
    LF-PIDNN & 0.3649  & 0.3454 & 0 & 0 & 64.35\% \\
    HF-DNN & 0.0275 & 1.9189  &  10 & $10\times 5$ &  98.04\% \\
    MF-PIDNN & 0.17 & 0.9542  & 10  & $10\times 5$ & 1.52\% \\ \hline
    \end{tabular}
\end{table}

\begin{figure}[ht!]
    \centering
    \subfigure[]{
    \includegraphics[width=0.48\textwidth]{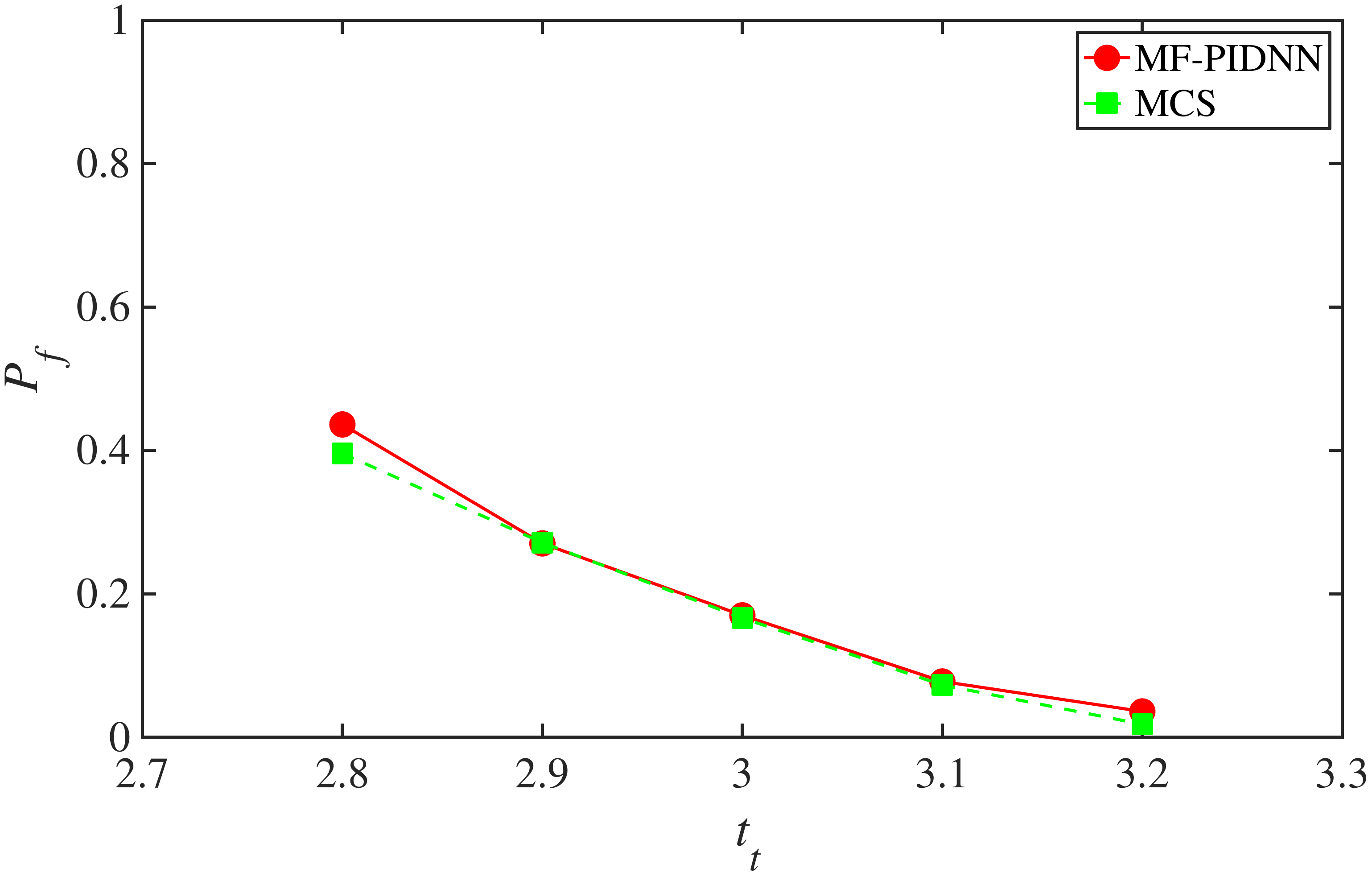}}
    \subfigure[]{
    \includegraphics[width=0.48\textwidth]{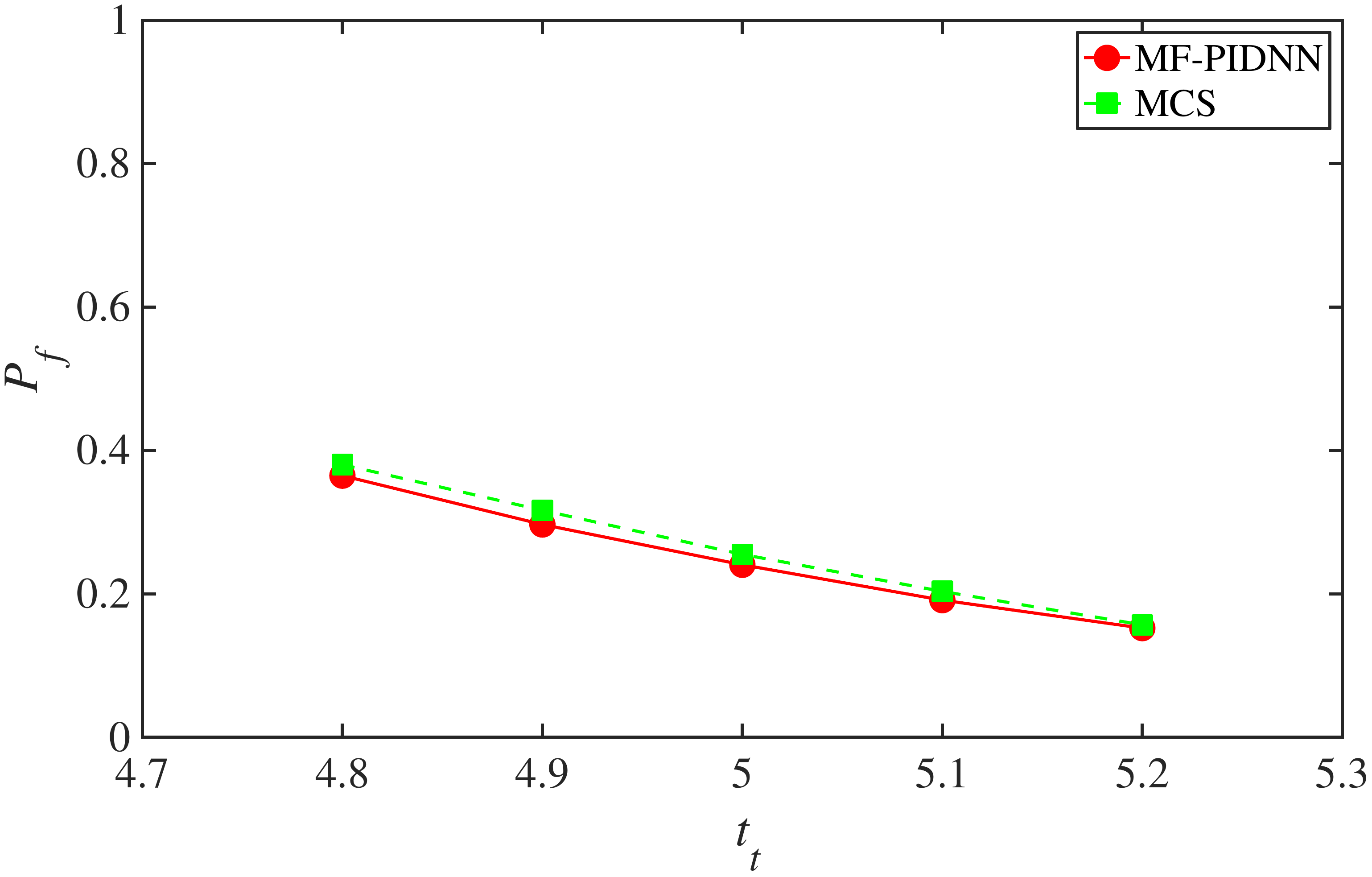}}
    \subfigure[]{
    \includegraphics[width=0.48\textwidth]{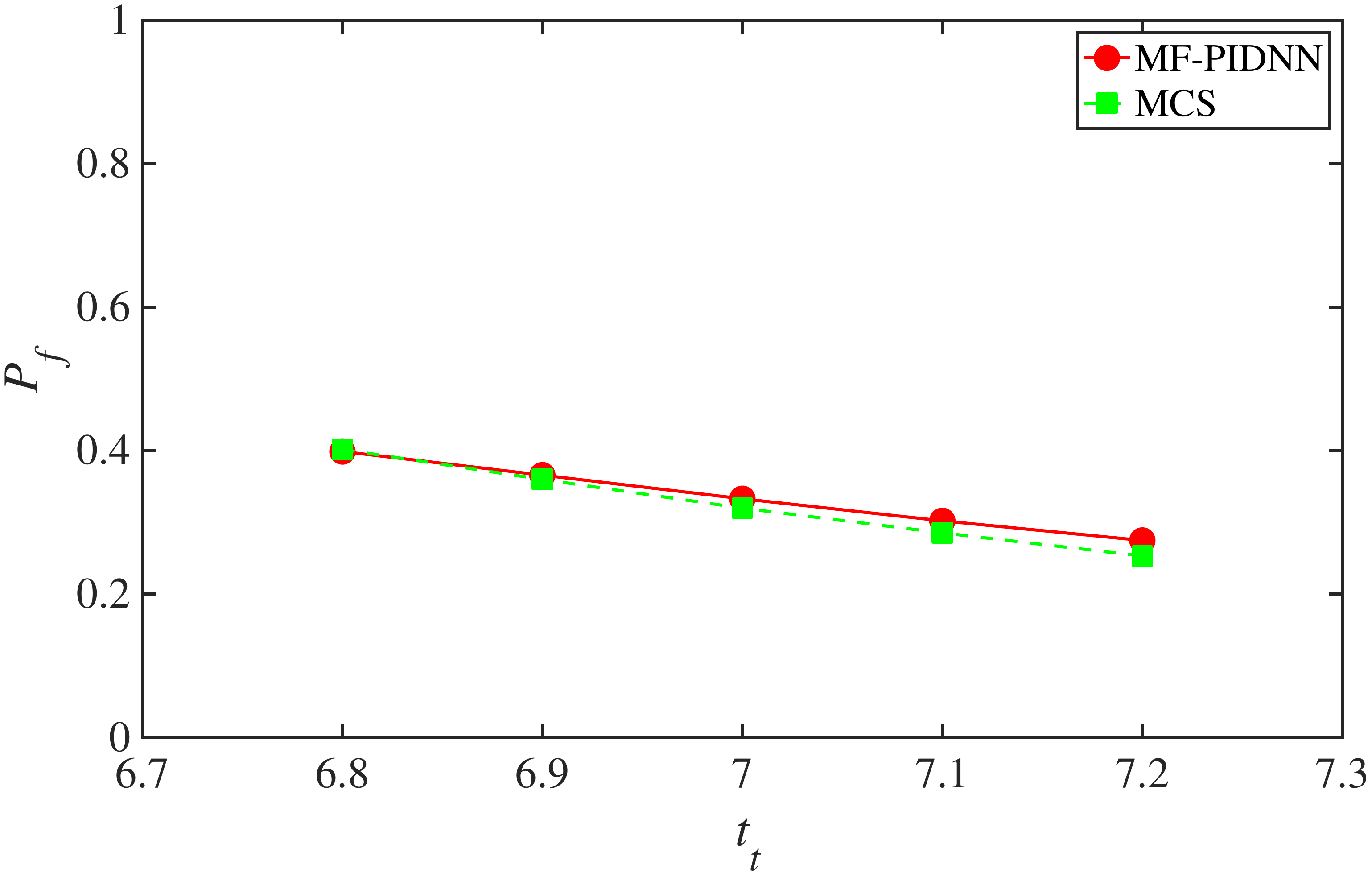}}
    \subfigure[]{
    \includegraphics[width=0.48\textwidth]{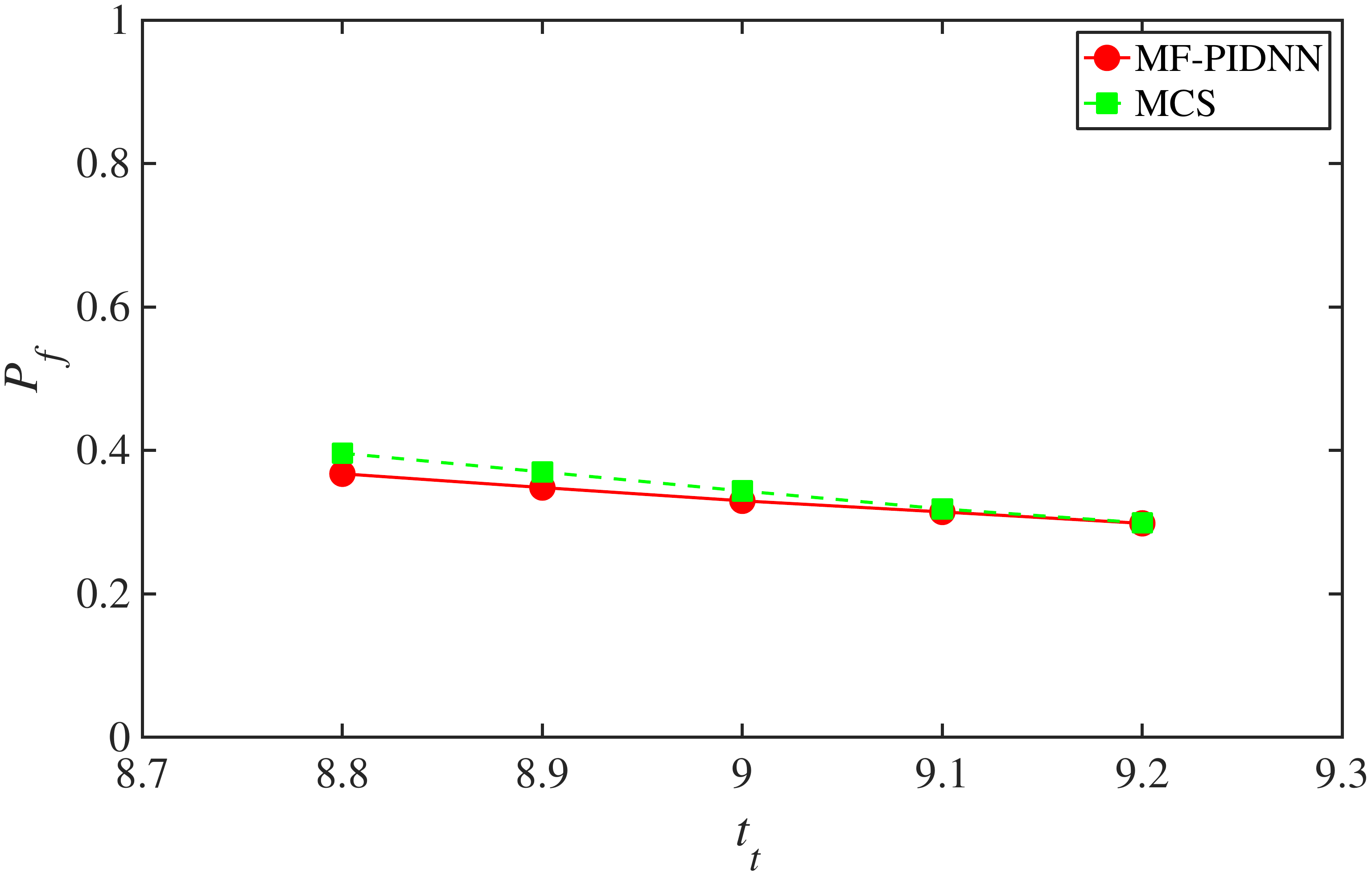}}
    \caption{MF-PIDNN and MCS predicted results at different time-instants. The threshold $e_{3,0}$ for these four cases are set at (a) 0.40, (b) 0.575, (c) 0.70 and (d) 0.78.}
    \label{fig:eg4_time_var}
\end{figure}

Finally, the trained MF-PIDNN is used to compute the
probability of failure at different time instants.
The corresponding results are illustrated in 
\autoref{fig:eg4_time_var}. 
To be specific, probability of failures around $t=3, 5, 7$ and 9 are presented. The threshold $e_{3,0}$
for the four cases are set to be 0.40, 0.575, 0.70 and 0.78.
MF-PIDNN for all the four cases is found to yield 
reasonably accurate results. Do note that high-fidelity data was only available at five equidistant time-instants between $t = 4.0$ and $t=7.0$. The fact that the proposed approach yields reasonable results outside this domain illustrates
the extrapolability of the proposed approach.
This capability of the MF-PIDNN is because of the 
fact that some physics is learnt (and retained) from the low-fidelity data.

\section{Conclusions}\label{sec:conclusions}
In this paper, a multi-fidelity physics informed deep 
neural network (MF-PIDNN) is presented.
The proposed approach is ideally suited for problems
where the physics of the problem is known
in an approximate sense (low-fidelity physics) and 
only a few high-fidelity data is available. 
MF-PIDNN blends the concepts of physics-informed and data-driven deep learning; 
the primary idea is to first train a low-fidelity
deep learning model based on the available approximate physics and then use transfer learning to update the model based on the high-fidelity data.
With this, MF-PIDNN is able to extract useful information from both the low-fidelity physics and high-fidelity data. There are two distinct advantages of MF-PIDNN. 
First, the low-fidelity model is directly trained from the physics of the problem and hence, no low-fidelity data is needed in this framework.
Second, because of the physics-informed framework
within MF-PIDNN, the proposed approach is able to capture some of the physical laws that are present
in the approximate model. As a result, it provides
reasonable predictions even in zones with no-data.

The proposed approach is used for solving benchmark
reliability analysis problems from the literature.
For all the problems, the proposed approach is able 
to correctly predict the probability of failure 
and the reliability index of the system.
To illustrate the advantage of the proposed approach, results obtained are compared with those
obtained from only the high-fidelity data-driven model and low-fidelity physics-driven model.
The proposed approach is found to outperform both these approaches.
Case studies are also presented to illustrate different features of MF-PIDNN.

Despite the several advantages of the MF-PIDNN, certain aspects can be further enhanced.
For example, during updating the model using transfer learning, mean-squared loss-function with no regularization has been used.
This can lead to over-fitting. One future direction
is to study the effect of regularization on the results.
Second, the number of tunable parameters during transfer learning are selected manually in this study.
Automating the transfer learning step will be hugely
beneficial.
Third, the network architecture and the activation
functions in this study are manually provided.
Automating this will also be beneficial.
In future, some of these aspects will be investigated.
\section*{Acknowledgements}
The author would like to thank Soumya Chakraborty for proof-reading this article and Somdatta Goswami,
Tanmoy Chatterjee and Rajdip Nayek for the useful discussions during the preparation of this paper. The \texttt{TensorFlow} codes were run on Google Colab service.



\end{document}